\documentclass[letterpaper, 10 pt, conference]{ieeeconf}  

\IEEEoverridecommandlockouts                              

\usepackage{amssymb}  
\usepackage[T1]{fontenc}
\usepackage{todonotes}
\usepackage{graphicx}
\usepackage{subcaption}
\usepackage{tabularx}
\usepackage{array}
\usepackage{amsmath}
\usepackage{booktabs}
\usepackage{multirow}
\usepackage{booktabs}
\usepackage{adjustbox}

\title{\LARGE \bf
Data-centric Design of Learning-based Surgical Gaze  Perception Models in Multi-Task Simulation
}

\author{Yizhou Li$^{1}$, Shuyuan Yang$^{1}$, Jiaji Su$^{1}$, Zonghe Chua$^{1}$%
\thanks{This work was supported NIH Award Number 1R21EB036161-01.}
\thanks{$^1$Y. L., S. Y., J. S., and Z. C. are with the Department of Electrical, Computer, and Systems Engineering, Case Western Reserve University, Cleveland, OH 44106 USA.
        \{yxl3527, sxy841, jxs1778, zxc703\}@case.edu.}%
}

\begin{document}

\maketitle
\thispagestyle{empty}
\pagestyle{empty}

\begin{abstract}

In robot-assisted minimally invasive surgery (RMIS), reduced haptic and depth cues increase reliance on expert visual perception, motivating gaze-guided training and learning-based surgical perception models. However, operative expert gaze is costly to collect, and it remains unclear how the source of gaze supervision, both expertise level (intermediate vs novice) and perceptual modality (active execution vs passive viewing), shapes what attention models learn. We introduce a paired active–passive, multi-task surgical gaze dataset collected on the da Vinci SimNow simulator across four drills. Active gaze was recorded during task execution using a VR headset with eye tracking, and the corresponding videos were reused as stimuli to collect passive gaze from observers, enabling controlled same-video comparisons. We quantify skill- and modality-dependent differences in gaze organization and evaluate the substitutability of passive gaze for operative supervision using fixation-density overlap analyses and single-frame saliency modeling. Across settings, MSI-Net produced stable, interpretable predictions, whereas SalGAN was unstable and often poorly aligned with human fixations. Models trained on intermediate passive gaze recovered a substantial portion of intermediate active attention, but with predictable degradation, and transfer was asymmetric between active and passive targets. Notably, novice passive labels approximated intermediate-passive targets with limited loss on higher-quality demonstrations, suggesting a practical path for scalable, crowd-sourced gaze supervision in surgical coaching and perception modeling.

\end{abstract}

\section{Introduction}

In robot-assisted minimally invasive surgery (RMIS), diminished haptic cues and depth perception make expert visual perception highly critical for achieving high performance \cite{lazar_seventh_2019, chua_task_2020}. Such cues can inform both clinical decision making and task execution, e.g., through identifying abnormal anatomy, and safe tissue dissection planes, respectively. 

Prior works have shown that expert and novice surgeons exhibit distinct gaze behaviors during minimally invasive procedures and training tasks \cite{wilson_psychomotor_2010,khan2012analysis}. These reflect differences in perceptual--cognitive control rather than incidental looking. Notably, experts tend to adopt a stable, target-focused strategy (often described as target-locking or quiet-eye-like behavior), whereas novices more frequently switch between tool and target to monitor action consequences and coordinate visuomotor control \cite{vine2012cheating,vine2013gaze}. 

These insights have led to engineered instructional systems and expert-designed pedagogical approaches to use quiet eye strategies for skill improvement. Guiding novices visually using visual overlays to adopt this gaze strategy improved performance and supports retention and transferred to more complex manual laparoscopic skills \cite{vine2013gaze}. Similar evidence exists in robotic surgery simulation, where gaze-augmented in-person coaching helped learners internalize target-centered visual strategies and can improve robustness under multitasking or stress \cite{melnyk2021see}. 

The above results motivate the broader question of whether expert surgical gaze encodes reusable perceptual strategies that can be transferred not only between humans, but also to learning-based models to enhance task performance. In other domains, such as autonomous driving \cite{liu2019utilizing}, and medical imaging \cite{zhu2022gaze, ma2023eye, antunes2023generating}, gaze data has been used to shape model attention and improve generalization. However, developing learning-based surgical perception models requires addressing unique data-centric engineering challenges: that acquiring expert surgical perception data at scale is difficult and expensive to obtain, and that expert perception models might need to generalize to contexts encountered by non-expert users and systems.

One potential approach to these issues is crowd-sourcing. This approach has proven successful for video-based surgical skill evaluations, in which non-expert crowd-sourced ratings converged reliably towards expert ratings \cite{lendvay_crowdsourcing_2015}. However, unlike with surgical skill evaluation, surgical perception is usually coupled to sensorimotor action. During active task execution, gaze is coupled to action planning, control, and error monitoring. In contrast, passive viewing may reflect a different set of cognitive processes, such as comprehension and post-hoc interpretation. Khan et al.\ used a gaze-overlap analysis to compare operative gaze with gaze recorded during video viewing, finding that experts' viewing gaze overlaps with their own operative gaze more than junior residents' viewing gaze, particularly during more complex steps \cite{khan2012analysis}. Overall, it remains an open question whether learning-based perceptual models capturing core features of expert surgical gaze can be trained on these different data sources (expert vs. novice) and perceptual modes (passive vs. active observation).

In this work, we present a systematic evaluation of how two critical dimensions of source data, expertise level, and perceptual modality, affect the ability of learning-based surgical perception models to replicate expert surgical gaze. We achieved this by conducting a novel human subject study on a surgical simulator with a custom gaze measurement pipeline. The generated data was used to train several learning-based surgical perception model variants which we analyzed using standard saliency map quantification metrics.

\section{Dataset Collection Protocol}
\subsection{Overview}
We conducted a human subjects experiment on a da Vinci Surgical Simulator (with SimNow) to collect complementary telesurgical gaze data that varied along two dimensions: skill level, and gaze modality. These two independent variables were crossed to produce four experimental conditions.

Skill level was divided into novice and intermediate. Novice participants were defined as having \emph{less than 1 hour} of self-reported prior experience with the da Vinci Surgical System. Intermediate demonstrators were defined using a reproducible performance criterion on the simulator: achieving an overall SimNow score $\ge 95$ on each of the four selected drills (Task~A--D; see Sec.~\ref{sec:exp_protocol}) for three consecutive attempts per drill. SimNow scores are construct-validated, and our criteria were set higher than proficiency thresholds seen in other studies \cite{turbati2023robotic,gomes2025construct,applegarth2026evaluating}.

Observation modality was divided into \emph{active} gaze, which was measured as participants performed four surgical simulation tasks, and \emph{passive} gaze, which was measured as participants observed video recordings of the surgical simulation tasks performed in the active gaze condition.

\begin{figure*}[t]
\vspace{0.5em}
  \centering
  \begin{minipage}[t]{5in}
    \vspace{0pt}
    \centering
    \includegraphics[width=\textwidth]{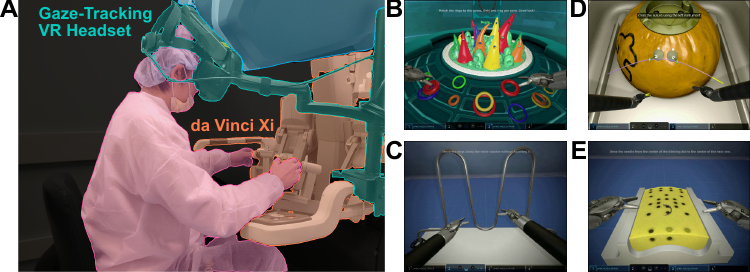}
  \end{minipage}\hfill
  \begin{minipage}[t]{1.95in}
    \vspace{0pt}
    \caption{(A) Active gaze collection experimental setup in which video output from the da Vinci surgeon console is piped into a VR headset with native gaze-tracking functionality. da Vinci Surgical Simulation SimNow tasks: (B) Sea Spikes, (C) Ring Rollercoaster~1, (D) Knot-tying, and (E) Big Dipper Needle Driver.}
    \label{fig:setup}
  \end{minipage}
\end{figure*}

\subsection{Participants}
\label{sec:participants}

A total of 12 first-year medical students took part of an IRB-approved study design, and were assigned to the novice group. Due to the practical difficulty of recruiting experienced surgeons for this study, we did not include a board-certified expert group. 
Pilot demonstration data were collected at the intermediate level. This comprised three trained members of the research team (non-surgeons) who practiced on SimNow over the course of several weeks until they met our specified criterion. Accordingly, we explicitly treated this group as \emph{intermediates} rather than \emph{experts} throughout the paper.

\subsection{Active Gaze Experiment}

\subsubsection{Hardware Setup}
We collected active eye-tracking data on the da Vinci SimNow Skills Simulator, a VR training module for the da Vinci Surgical System (Fig.~\ref{fig:setup}A). The active acquisition setup integrated a da Vinci Xi surgeon console running SimNow with a Varjo Aero head-mounted display (HMD), which has native eye-tracking capability. The stereo view output from the simulator was captured from the console at $1280\times 1024$ resolution per eye and 60~fps, streamed to the headset. This stereo video stream was recorded as a side-by-side stereo video at $2560\times 1024$ and 60~fps. Eye-tracking samples were recorded via the Varjo SDK at 200~Hz.

For each trial, we recorded (i) the time-stamped video stream, (ii) time-stamped gaze data, and (iii) SimNow performance reports including overall score, completion time, nominal score, and task-specific penalty counts. The overall score is the nominal score minus the weighted penalties.


\subsubsection{Experiment Protocol}
\label{sec:exp_protocol}
Novice participants first consented and then familiarized with the system. They completed the built-in SimNow introductory drill \texttt{Camera~0}, which introduced them to manipulator and endoscope movement, and tool clutching. Next, each novice watched a knot-tying demonstration video once \emph{without} additional instructions, to establish a shared reference of task completion. 

In the main experiment, participants were ask to complete four simulator tasks three times each in randomized order-- \textbf{Task A:} Sea Spikes~1, \textbf{Task B:} Ring Rollercoaster~1, \textbf{Task C:} Knot Tying, and \textbf{Task D:} Big Dipper needle driving (Fig.\,\ref{fig:setup}B--D, respectively). These drills were chosen to cover complementary psychomotor and bimanual manipulation skills relevant to robotic surgery training, while also spanning diverse failure/penalty modes (e.g., collisions, excessive force, drops, needle--tissue force events).

\subsection{Passive Gaze Experiment}
\label{sec:passive}
\subsubsection{Hardware Setup}
Recordings of the active gaze trials were presented as task videos on a 1920$\times$1080 monitor at a viewing distance of 60~cm. Although the recorded stimuli are side-by-side (SBS) stereo videos, passive observers viewed only the left-eye view (1280$\times$1024) in full-screen. Gaze data were recorded using a Gazepoint GP3 HD eye tracker at 150~Hz. For each passive trial, we recorded the observer’s gaze as normalized point-of-gaze coordinates along with per-sample timestamps

\subsubsection{Experiment Protocol}
\label{sec:exp_protocol}

Novice participants from the active gaze experiment were invited back for a follow-up session. The intermediate demonstrators from the study team likewise participated in this phase of the experiment to provide intermediate passive observation data. 

Each novice participant watched seven videos in a fixed alternating order: I--N--I--N--I--N--I, including four intermediate-source segments (one each for Tasks A through D) and three novice-source segments (one each for Tasks A through C). This alternating order was chosen to mitigate visual fatigue because novice-source attempts tend to be longer in duration. To direct their attention purposefully, participants were told that their goal was to view the videos and at the end of each, rate performance using a rubric. They took 5-minute breaks after every three videos. 

Due to the small number of Intermediate observers, they were assigned an additional 15 segments(including both novice-source and intermediate-source segments) beyond the novice schedule to increase passive examples at the intermediate level. Intermediates did not therefore strictly view the videos in I--N--I--N--I--N--I order. Instead, the additional segments were arranged to interleave novice and intermediate-source segments as much as possible. Participants were not allowed to pause or replay videos. 

After each viewed attempt, observers completed a short task-specific questionnaire to rate the performer’s perceived performance.

\subsection{Allocation of Active Gaze Demonstrations for Observation}
Intermediates and novices were assigned to watch both intermediate and novice demonstrations. To reduce the onset of cognitive fatigue and maintain a reasonable single-session experimental duration, we selectively sampled from our active gaze trials to generate subsets for each task, and expertise level, in a manner that balanced diversity across different demonstrators and task errors (as measured via SimNow penalty scores). 

To construct a subset for each task for a particular expertise level, all active demonstrations of the task were first ranked by a performance index
\begin{equation}
    \mathrm{PerfIndex}(n) = z\big(s(n)\big) - z\big(p(n)\big) - z\big(t(n)\big),
\end{equation} 
where $s(n)$ is score, $t(n)$ is time, and $p(n)$ is the penalty score for the $n^\text{th}$ demonstration, and $z(\bullet)$ denotes per-task z-normalization 

Novice demonstrations were ranked in ascending order of performance (i.e. worse performance first) and vice versa for intermediate demonstrations. The trial number across the entire duration of a participant's active experiments trials, was used to break ties, with, earlier tasks ranked higher, given that these have higher observed movement and strategy variability, and thus greater diversity. According to this ranked list, one demonstration per participant was selected. This led to each task A to C having 11 demonstrations each in the novice sets, and each task A to D having 3 demonstrations each, in the intermediate sets.

During the experiment, participants then viewed one video from each of these subsets. To ensure consistency across viewing conditions, no participant viewed their own demonstrations. As noted in Sec.\,\ref{sec:exp_protocol}, intermediates were also assigned to view fifteen more novice examples sampled from each novice task subset.

\subsection{Data Post-Processing}
For both active and passive datasets, we detected fixations using the I-DT (dispersion-threshold) \cite{salvucci2000identifying}  algorithm. 
A gaze segment spanning the i$^\text{th}$ to the j$^\text{th}$ frame is a fixation if
(i) its duration exceeds a minimum threshold
$T_{\min}=0.10\text{s}$, and (ii) its spatial dispersion is below a threshold $D_{\max}=50\text{px}$, i.e. 
\begin{equation*}
(\Delta t_j - \Delta t_i)\ge T_{\min}
\quad\text{and}\quad
D(i,j)\le D_{\max},
\label{eq:idt-criteria}
\end{equation*}
where dispersion of the segment is
\begin{equation}
D(i,j) =
\left(\max_{k\in[i,j]} x_k - \min_{k\in[i,j]} x_k\right)
+
\left(\max_{k\in[i,j]} y_k - \min_{k\in[i,j]} y_k\right).
\label{eq:dispersion}
\end{equation}
Consequently, $\{[s_m,e_m]\}_{m=1}^{M}$ denote the detected fixation segments. All samples not assigned to any fixation segment are treated as non-fixation.

We converted the gaze point $(x_i,y_i)$ into a frame-by-frame 2D Gaussian heatmap.
For $(u,v)$ index pixels on the gaze-map grid, the Gaussian density was
\begin{equation}
    G_i(u,v) = \exp\left(-\frac{(u-x_i')^2 + (v-y_i')^2}{2\sigma^2}\right),
\end{equation}
where $(x_i',y_i')$ is the gaze point mapped into pixel space. We thresholded the Gaussian at radius $r = t\cdot\sigma$ and set values outside the radius to zero. We used $\sigma = 5.0$ pixels and $t = 3.0$.

To reduce storage and training costs, we downsampled frames and labels.
Specifically, we resized each stimulus frame from $1280\times 1024$ to $640\times 512$ for model input, and generated the gaze maps at $160\times 128$ resolution (i.e., 4$\times$ downsampled relative to $640\times 512$, and 8$\times$ relative to $1280\times 1024$).

\subsection{Validation Metrics}
\label{sec:Validation_metrics}

The resultant gaze data from each experiment were analyzed using several gaze metrics consistent with prior gaze-base skill assessment formulations~\cite{li2023eye}. To account for the significantly longer durations of novice demonstrations, all metrics were duration normalized.

\subsubsection{Fixation count and fixation time.}
The number of fixations is $N_{\mathrm{fix}} = M$.
The fixation time is
\begin{equation}
T_{\mathrm{fix}}
= \sum_{m=1}^{M}\left(\Delta t_{e_m}-\Delta t_{s_m}\right),
\label{eq:fix-time}
\end{equation}
and we report summary statistics (mean/median/std) over the fixation duration set
$\left\{\Delta t_{e_m}-\Delta t_{s_m}\right\}_{m=1}^{M}$.

\subsubsection{Fixation-to-non-fixation ratio.}
Following common practice, we compute
\begin{equation}
R = \frac{T_{\mathrm{fix}}}{T_{\mathrm{total}} - T_{\mathrm{fix}}}.
\label{eq:fix-nonfix-ratio}
\end{equation}
When $(T_{\mathrm{total}} - T_{\mathrm{fix}})$ is numerically close to zero, $R$ is treated as undefined.

\subsubsection{Scanpath rate (fixation-center based).}
For each fixation segment $m$, we compute its center as the mean gaze position within the segment:
\begin{equation}
\mathbf{c}_m =
\left(
\frac{1}{e_m-s_m+1}\sum_{t=s_m}^{e_m}x_t,\;
\frac{1}{e_m-s_m+1}\sum_{t=s_m}^{e_m}y_t
\right).
\label{eq:fix-center}
\end{equation}
We define the scanpath length as the sum of Euclidean distances between consecutive fixation centers:
\begin{equation}
L_{\mathrm{scan}} = \sum_{m=2}^{M}\left\|\mathbf{c}_m - \mathbf{c}_{m-1}\right\|_2.
\label{eq:scanpath}
\end{equation}
This was then normalized by demonstration duration to obtain an average scanpath speed.

\subsection{Convex Hull}
Convex hulls were computed with the Graham scan algorithm~\cite{graham1972efficient}. This method deterministically constructs the minimal enclosing convex polygon from a set of planar points by iteratively enforcing counter-clockwise orientation constraints. The resulting hull provides an order-invariant and geometrically well-posed estimate of the spatial support of gaze samples. Hull area was subsequently used as a summary measure of visual exploration extent, with smaller areas indicating more spatially focused scanning behavior.

\subsection{FDM}
In our implementation, the fixation density map (FDM) is constructed from discrete fixation events: for each fixation, we take the event center as the median gaze location within the fixation segment, accumulate the fixation duration as a weight at the corresponding pixel, and apply Gaussian smoothing with $\sigma=30$ pixels. The resulting map is then normalized by its sum to obtain a probability distribution $p$. The similarity metric \textbf{FDM-SIM} is defined as the histogram intersection between two probability FDMs,
$\mathrm{SIM}(p,q)=\sum_{x,y}\min\!\big(p(x,y),q(x,y)\big)$.
The correlation metric \textbf{FDM-CC} is defined as the Pearson correlation coefficient between the flattened unnormalized FDM heatmaps,
$\mathrm{CC}(hm_A,hm_P)=\mathrm{corr}\!\big(\mathrm{vec}(hm_A),\mathrm{vec}(hm_P)\big)$.

\section{Computational Approach}

\subsection{Computational Gaze Models}

For our study, we chose to model contextual scene perception, and hence selected several frame-wise spatial attention prediction models over spatio-temporal models, which are able to model transient gaze dynamics, but had more complex architectures that could be easily overfit. To cover complementary modeling paradigms and surface trade-offs, we compared three representative single-frame saliency approaches: 
MSI-Net~\cite{kroner2020contextual}, a supervised CNN-based contextual encoder–decoder, a standard and deployment-friendly deep saliency predictor, and SalGAN~\cite{pan2017salgan} (a GAN-based saliency model) as a generative alternative that encourages saliency-map realism beyond pixel-wise supervision.

\subsection{Expertise- and Modality-Conditioned Dataset Splits and Resultant Data-centric Model Variants}

We split the dataset into training, validation, and test sets across tasks. We adopted a \textit{expertise-} and \textit{modality-conditioned} design approach, and created four distinct training dataset: Intermediate-Active (IA), Intermediate-Passive (IP), Novice-Active (NA), and Novice-Passive (NP). IA data consisted of intermediate active gaze labels only, while IP consisted of intermediate passive gaze labels, and their corresponding demonstrations, with NA and NP similarly defined for novice gaze labels. Critically, active datasets IA and NA, had corresponding demonstrations only at that particular skill level (i.e. IA only had intermediate demonstrations, and NA only novice ones), while passive datasets IP and NP contained demonstrations from both expertise levels. The total number of gaze demonstrations for each task and each expertise level are summarized in Table\,\ref{tab:dataset_split}. Each training dataset was used to train a expertise-modality variant of each computational gaze-model resulting in four variants  IA, IP, NA and NP for each learning-based models.

Validation and test sets include covered all available tasks and skill groups. For analysese, the test set could be subdivided into smaller subsets according to task, gaze modality (active/passive), gaze source (intermediate/novice) and demonstration expertise level (intermediate/novice). All data derived from the same trial, including the operator's active gaze annotation and any passive gaze annotations collected from observers watching the same trial video, are assigned to the same split to prevent leakage of visual content across sets.

\begin{table}[t]
\label{tab:dataset_split}
\vspace{0.5em}
\caption{Distribution of Demonstrations Across Splits}
\begin{tabular*}{\columnwidth}{@{\extracolsep{\fill}}c cc cc cc}
\toprule
\multirow{2}{*}{\textbf{Task}} 
& \multicolumn{2}{c}{\textbf{Train}} 
& \multicolumn{2}{c}{\textbf{Val}} 
& \multicolumn{2}{c}{\textbf{Test}} \\
\cmidrule(lr){2-3} \cmidrule(lr){4-5} \cmidrule(lr){6-7}
 & \textbf{I} & \textbf{N} 
 & \textbf{I} & \textbf{N} 
 & \textbf{I} & \textbf{N} \\
\midrule
Task A & 5(3) & 23(8) & 2(9) & 5(4) & 2(5) & 5(3) \\
Task B & 5(3) & 25(9) & 2(6) & 4(6) & 2(4) & 4(4) \\
Task C & 6(4) & 41(17) & 2(6) & 5(3) & 2(4) & 5(4) \\
Task D & 5(4) & 0(0) & 2(6) & 0(0) & 2(4) & 0(0) \\
\bottomrule
\multicolumn{7}{l}{() denotes the subset of passive gaze demonstrations available.}
\end{tabular*}
\label{tab:dataset_split}
\end{table}

\begin{table*}[!t]
\centering
\vspace{0.5em}
\caption{Performance Summary of da Vinci SimNow tasks in Active dataset}
\label{tab:active_coverage}
\begin{tabularx}{\textwidth}{
    >{\centering\arraybackslash}p{0.4cm}   
    >{\centering\arraybackslash}p{1.4cm}   
    >{\centering\arraybackslash}p{2.5cm}   
    >{\centering\arraybackslash}p{4.4cm}   
    >{\centering\arraybackslash}p{4.2cm}   
    >{\centering\arraybackslash}p{2.4cm}   
}
\hline
\textbf{Task} & 
\textbf{Group} & 
\textbf{Score} (mean$\pm$std) & 
\textbf{Economy of Motion} (cm, mean$\pm$std) & 
\textbf{Completion Time} (s, mean$\pm$std) & 
\textbf{Complete rate. (\%)} 
\\
\hline
A & Novice       & 38.94 $\pm$ 29.77 & 463.32 $\pm$ 155.93 & 279.94 $\pm$ 92.01  & 100 \\
A & Intermediate & 96.78 $\pm$ 2.73  & 259.06 $\pm$ 19.38  & 93.28 $\pm$ 12.76   & 100 \\
B & Novice       & 29.26 $\pm$ 32.28 & 253.06 $\pm$ 88.74  & 305.68 $\pm$ 180.53 & 100 \\
B & Intermediate & 98.44 $\pm$ 1.13  & 132.90 $\pm$ 10.52  & 60.07 $\pm$ 15.60   & 100 \\
C & Novice       & 16.37 $\pm$ 26.78 & 259.02 $\pm$ 135.95 & 340.84 $\pm$ 171.40 & 69.7 \\
C & Intermediate & 76.22 $\pm$ 43.28 & 78.09 $\pm$ 19.84   & 190.06 $\pm$ 71.26  & 92.6 \\
D & Novice       & -- & -- & -- & -- \\
D & Intermediate & 94.40 $\pm$ 2.55  & 277.39 $\pm$ 60.65  & 242.10 $\pm$ 35.35  & 100 \\
\hline
\end{tabularx}
\end{table*}

\begin{figure*}[!t]
  \centering
  \includegraphics[width=\textwidth]{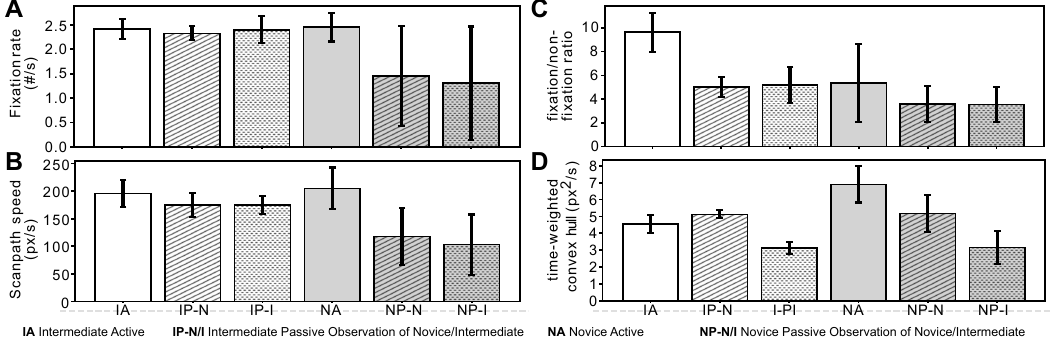}
  \caption{Gaze metrics for active and passive demonstrations. (A) Fixation rate, (B) scanpath speed, (C) fixation ration, (D) convex hull. All metrics are time normalized over the duration of each demonstration.}
  \label{fig:gaze_metrics}
\end{figure*}

\begin{figure}[t]
  \centering
  \includegraphics[width=\columnwidth]{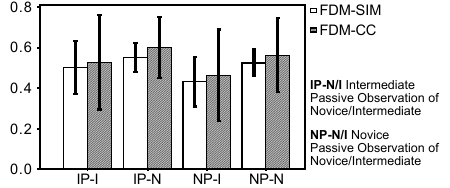}
  \caption{Gaze overlap metrics across observer--source skill combinations. Bars indicate mean values and error bars denote standard deviation.}
  \label{fig:fdm_sim_skill}
\end{figure}

\begin{figure}[!t]
  \centering
  \includegraphics[width=\dimexpr\linewidth-5pt\relax]{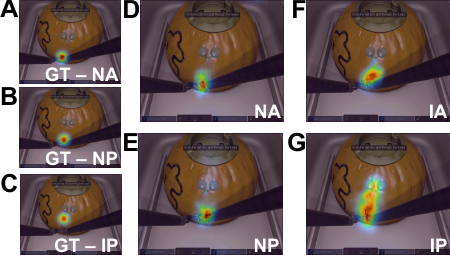}
  \caption{Selected heatmaps for ground truth (A) NA – novice active gaze, (B) NP – novice passive gaze, (C) IP – intermediate passive gaze, and MSI-Net model predictions trained on (D) NA (E) NP (F) IA  (G) IP predictions.}
  \label{fig:heatmaps}
\end{figure}


\subsection{Training Procedures}

MSI-Net was trained using the Adam optimizer~\cite{kingma2014adam} with batch size $8$, learning rate $1\times10^{-4}$, $\beta=(0.9,0.999)$, and weight decay $0$, for up to $80$ epochs. We minimized the Kullback--Leibler divergence between the ground-truth saliency map $g$ and the predicted saliency map $p$:
\begin{equation}
\mathcal{L}_{\mathrm{KLD}}(g\Vert p)=\sum_{i} g_i \log\left(\epsilon + \frac{g_i}{\epsilon + p_i}\right),
\label{eq:kld_def}
\end{equation}
where $g$ and $p$ are clamped to be non-negative and normalized to sum to $1$ per sample. We applied a ReduceLROnPlateau scheduler (factor $0.5$, patience $2$, minimum learning rate $1\times10^{-6}$) and early stopping (patience $8$) based on validation performance. Automatic mixed precision was disabled.

SalGAN was trained using Adam optimizers~\cite{kingma2014adam} for both the generator and discriminator with batch size $8$. The generator used learning rate $2\times10^{-5}$, $\beta_G=(0.9,0.999)$, and weight decay $0$, while the discriminator used learning rate $4\times10^{-4}$, $\beta_D=(0.5,0.999)$, and weight decay $0$. During adversarial training, the discriminator input was formed by concatenating the downsampled video frame $f$ and the saliency map along the channel dimension:
\begin{equation}
x_{\text{real}}=[f,\,S_{\text{gt}}],\quad
x_{\text{fake}}=[f,\,S_{\text{pred}}],
\end{equation}
and the discriminator was trained with a BCE-with-logits objective to distinguish $d_{\text{real}}$ from $d_{\text{fake}}$ with label smoothing for real targets set to $0.9$. The generator was trained to match the ground truth saliency map via a reconstruction term, and prevent the discriminator from distinguishing $d_{\text{real}}$ to $d_{\text{fake}}$:
\begin{equation}
\mathcal{L}_G = \mathcal{L}_{\mathrm{adv}} + \alpha \, \mathcal{L}_{\mathrm{recon}},
\end{equation}
where $\alpha=0.005$, $\mathcal{L}_{\mathrm{recon}}$ is a marked BCE computed at the label resolution. We used an alternating update schedule with $n_{\text{critic}}=5$, i.e., five discriminator updates per each generator update, with gradient clipping at a maximum norm of $1.0$ and a learning rate scheduler with factor $0.5$ and patience $1$.

\subsection{Validation Metrics}


\paragraph{Validation metrics.}
We assess model performance against the baseline by comparing the predicted saliency map to the ground-truth saliency map on a per-frame basis, and averaging results over valid frames only. For each frame, the model output is clamped to $[0,1]$ and resized to the ground-truth resolution via bilinear interpolation. Let $y \in \mathbb{R}^{H\times W}$ denote the ground-truth saliency map and $\hat{y} \in \mathbb{R}^{H\times W}$ the prediction at the same resolution. For distribution-based metrics, we normalize maps to sum to one:
$p = \frac{\max(y,0)}{\sum \max(y,0) + \epsilon}$ and
$q = \frac{\max(\hat{y},0)}{\sum \max(\hat{y},0) + \epsilon}$, with a small $\epsilon$ for numerical stability.

We evaluate saliency predictions against the baseline using KLD, CC, SIM, and NSS, computed per valid frame and averaged across frames and videos. \textbf{KLD} is reported using the same definition and normalization as in Eq.~(\ref{eq:kld_def}).

\textbf{CC} is computed as the Pearson correlation coefficient between the flattened maps (without sum-to-one normalization):
\begin{equation}
\mathrm{CC}(y,\hat{y}) = \frac{1}{N}\sum_{i=1}^{N}\tilde{y}_i\,\tilde{\hat{y}}_i,
\quad
\tilde{y}=\frac{y-\mu_y}{\sigma_y+\epsilon},\;
\tilde{\hat{y}}=\frac{\hat{y}-\mu_{\hat{y}}}{\sigma_{\hat{y}}+\epsilon},
\end{equation}
where $N=HW$, and higher values indicate better spatial agreement.

\textbf{SIM} is computed as histogram intersection between the normalized distributions:
\begin{equation}
\mathrm{SIM}(p,q) = \sum_{x,y}\min\!\big(p(x,y),\,q(x,y)\big),
\end{equation}
with higher values indicating better overlap.

\textbf{NSS} is computed by z-scoring the predicted map and sampling it at the ground-truth peak location. Specifically, let $(x^\ast,y^\ast)=\arg\max_{x,y} y(x,y)$, and
$z=\frac{\hat{y}-\mu_{\hat{y}}}{\sigma_{\hat{y}}}$ (computed with $\mathrm{ddof}=0$). Then
\begin{equation}
\mathrm{NSS}(y,\hat{y}) = z(x^\ast,y^\ast),
\end{equation}
where higher values indicate better alignment between the prediction and the ground-truth fixation location.

\section{Results and Discussion}

\subsection{Active Gaze Dataset}

Among the recruited novices, 11 completed the full active-task battery, while one novice did not pass the training process and thus contributed only partial recordings; We dropped this user's data in statistical comparisons and model training. 

Performance results on the SimNow tasks are summarized in Table\,\ref{tab:active_coverage}. The overall performance of intermediates compared to novices were higher for all tasks, which confirms the skill difference between the two groups. Intermediates achieved an average SimNow score of 91.5 across tasks A--D compared to novices' average score of 28.2 across tasks A--C. Completion times were also shorter for intermediates, averaging 114.5s versus 308.8s for novices, for tasks A--C.

We observed clear skill-dependent differences in gaze behavior that were consistent with prior literature, thus supporting the construct validity of our experiment. Although fixation rates were comparable between novice and intermediates (Fig.\,\ref{fig:gaze_metrics}A - IA vs. NA), the fixation/non-fixation ratio was noticeably greater for intermediates compared to novices, suggesting that intermediates demonstrated more fixation behavior compared to novices (Fig.\,\ref{fig:gaze_metrics}C - IA vs. NA). Similarly Tien et al. reported that expert surgeons exhibited significantly higher fixation frequency and dwell time on task-relevant regions, whereas less-experienced surgeons distribute gaze more broadly across non-operative areas \cite{tien_differences_2015}.

Intermediate users exhibited greater scanpath speeds compared to novices, suggesting more frequent scanning behavior between fixation points (Fig.,\ref{fig:gaze_metrics}B – IA vs. NA). Their smaller convex hull area than novices indicated more spatially concentrated gaze patterns (Fig.,\ref{fig:gaze_metrics}D – IA vs. NA). These results present agreement with the observations of Fichtel et al., in which expert surgeons demonstrated reduced dwell time within areas of interest, reflecting a more flexible and strategically distributed visual sampling strategy rather than indiscriminate exploration \cite{fichtel_eye_2019}.

\subsection{Passive Gaze Dataset}

Intermediate passive observation displayed similar aggregate patterns to intermediate active observations as measured by fixation rate, scanpath speeds and convex hull area (Fig.\,\ref{fig:gaze_metrics}A, B, and D, IP-N and IP-I). However, novice passive observation fixation rate, scanpath speeds, and convex hull area were noticeably lower than both intermediate and novice observations (NP-N and NP-I). This suggests that intermediates were likely displaying scanning behaviors during passive observation, while novices were more intently focused. This is supported by the heatmaps shown in Fig.\,\ref{fig:heatmaps}A–C, where the intermediate passive gaze in \ref{fig:heatmaps}C fixates on the towers where the knot is tied as opposed to only fixating on the tools like in \ref{fig:heatmaps}A and B. For fixation ratio, both intermediate and novice passive observations showed lower ratios compared to intermediate active observations, and were more similar to novice active observations in this regard. 

Using the FDM metric to compare passive observation to active observations as in Fig.\,\ref{fig:fdm_sim_skill} shows that passive observation gaze maps overlapped similarly for their corresponding source, i.e. passive intermediate gaze and novice gaze, had similar FDM when compared against intermediate and novice active gaze maps, respectively. Notably, passive intermediate gaze also showed high overlap, suggesting that passive intermediate gaze might possibly exhibit coverage over active novice gaze. This was not observed in the case of passive novice gaze versus active intermediate gaze.

\begin{table}[t]
\centering
\vspace{0.5em}
\caption{Average saliency metrics for each setting (MSINET and SalGAN).}
\label{tab:setting_metrics_msinet_salgan}

\setlength{\tabcolsep}{4pt}
\renewcommand{\arraystretch}{0.92}

\begin{adjustbox}{max width=\textwidth}
\begin{tabular}{llrrrr}
\toprule
\textbf{Model} & \textbf{Train $\rightarrow$ Test} & \textbf{CC} & \textbf{KLD} & \textbf{SIM} & \textbf{NSS} \\
\midrule
\multirow{16}{*}{\textbf{MSINET}}
 & IA $\rightarrow$ IA     & 0.5401 & 1.4984 & 0.4358 & 6.1425 \\
 & IA $\rightarrow$ IP-I & 0.3587 & 2.4482 & 0.3036 & 3.5676 \\
 & IA $\rightarrow$ IP-N & 0.2260 & 3.2161 & 0.1941 & 2.1270 \\
 & IP $\rightarrow$ IA     & 0.3724 & 2.1399 & 0.2820 & 3.5834 \\
 & IP $\rightarrow$ IP-I & 0.4461 & 1.8294 & 0.3222 & 4.5116 \\
 & IP $\rightarrow$ IP-N & 0.3652 & 2.2088 & 0.2601 & 3.6037 \\
 & NA $\rightarrow$ NA     & 0.3969 & 2.2200 & 0.3032 & 4.1477 \\
 & NA $\rightarrow$ IA     & 0.4249 & 1.9855 & 0.3375 & 4.4719 \\
 & NA $\rightarrow$ IP-I & 0.3142 & 2.5974 & 0.2596 & 3.0161 \\
 & NA $\rightarrow$ IP-N & 0.2703 & 2.8280 & 0.2193 & 2.5684 \\
 & NP $\rightarrow$ NA     & 0.2499 & 3.2225 & 0.1953 & 2.3648 \\
 & NP $\rightarrow$ IA     & 0.3800 & 2.2242 & 0.2914 & 3.7270 \\
 & NP $\rightarrow$ IP-I & 0.4255 & 1.9631 & 0.3184 & 4.3774 \\
 & NP $\rightarrow$ IP-N & 0.3254 & 2.3898 & 0.2399 & 3.1702 \\
 & NP $\rightarrow$ NP-I & 0.4764 & 1.9038 & 0.3518 & 5.1251 \\
 & NP $\rightarrow$ NP-N & 0.3199 & 2.4654 & 0.2317 & 3.1622 \\
\midrule
\multirow{16}{*}{\textbf{SalGAN}}
 & IA $\rightarrow$ IA     & 0.3079 & 3.3735 & 0.2564 & 2.7573 \\
 & IA $\rightarrow$ IP-I & 0.2639 & 3.9967 & 0.2229 & 2.3711 \\
 & IA $\rightarrow$ IP-N & 0.1872 & 5.0145 & 0.1680 & 1.7023 \\
 & IP $\rightarrow$ IA     & 0.1912 & 4.8024 & 0.1661 & 1.7189 \\
 & IP $\rightarrow$ IP-I & 0.1972 & 4.9207 & 0.1688 & 1.7630 \\
 & IP $\rightarrow$ IP-N & 0.1617 & 5.5731 & 0.1428 & 1.4587 \\
 & NA $\rightarrow$ NA     & -0.0198 & 4.0125 & 0.0388 & -0.3073 \\
 & NA $\rightarrow$ IA     & -0.0472 & 4.0004 & 0.0382 & -0.4486 \\
 & NA $\rightarrow$ IP-I & -0.0325 & 3.9837 & 0.0387 & -0.3041 \\
 & NA $\rightarrow$ IP-N & -0.0080 & 3.9446 & 0.0400 & -0.1014 \\
 & NP $\rightarrow$ NA     & 0.1264 & 6.4784 & 0.1176 & 1.1107 \\
 & NP $\rightarrow$ IA     & 0.2431 & 4.4925 & 0.2008 & 2.1846 \\
 & NP $\rightarrow$ IP-I & 0.2460 & 4.6036 & 0.2023 & 2.1953 \\
 & NP $\rightarrow$ IP-N & 0.1767 & 5.6346 & 0.1520 & 1.5575 \\
 & NP $\rightarrow$ NP-I & 0.3070 & 3.9384 & 0.2435 & 2.7992 \\
 & NP $\rightarrow$ NP-N & 0.1862 & 5.6263 & 0.1565 & 1.7116 \\
\bottomrule
\multicolumn{6}{l}{I - Intermediate, N - Novice, A - Active, P - Passive}
\end{tabular}
\end{adjustbox}
\end{table}


\subsection{Computational Model Comparisons}

Table\,\ref{tab:setting_metrics_msinet_salgan} summarizes the results for each variant of MSI-Net and SalGAN, indicating several consistent and practically important patterns in how passive and active gaze supervision transfer across expertise levels and demonstration sources. 

Across all evaluated conditions, a clear model-dependent performance stratification was observed. For MSI-Net, correlation and saliency alignment remain consistently meaningful, with CC ranging approximately from 0.23 to 0.54 and NSS from 2.13 to 6.14 across the full suite of supervision and test configurations. In contrast, SalGAN exhibits markedly weaker and less stable behavior, with CC spanning $-$0.05 to 0.31 and NSS frequently near or below zero ($-$0.45 to 2.80), including several settings in which predicted saliency is anti-correlated with human fixation density (e.g., NA→IA: CC = $-$0.047, NSS = $-$0.449). Qualitative results of the MSI-Net predictions are provided in Fig.\,\ref{fig:heatmaps}D–F.

These results indicate that, under the limited and heterogeneous surgical gaze data regime considered here, adversarial saliency learning was insufficiently constrained and fails to reliably capture human attention structure. Supervised learning with MSI-Net, by comparison, consistently yielded interpretable and quantitatively meaningful alignment. Accordingly, all subsequent analyses focus on MSI-Net to isolate data-centric effects from confounding model instability.

Considering passive-to-active transfer, MSI-Net trained on intermediate-passive labels achieved a CC of 0.372 and an NSS of 3.58 when evaluated against intermediate-active gaze (IP→IA), compared to the in-domain operative ceiling of CC = 0.540 and NSS = 6.14 for IA→IA. Thus, intermediate observational supervision recovers a substantial, but clearly incomplete, portion of the action-coupled intermediate attention signal. This similarity of IP and IA models can be partially observed in Figs.\,\ref{fig:heatmaps}F and IP, where the heatmaps for both the IA and IP models both overlap to some extent, both the towers and the tools. This was also seen for novice-passive gaze models (NP→IA) where. However, there was an asymmetry between passive and active attention. Models trained on active gaze generalize poorly to passive targets (e.g., IA→IP-I: CC = 0.359, NSS = 3.57).

When the intermediate-passive model predicted intermediate-passive gaze, performance was higher on intermediate demonstrations than on novice demonstrations: IP→IP-I attained CC = 0.446 and NSS = 4.51, whereas IP→IP-N dropped to CC = 0.365 and NSS = 3.60. This trend was not observed for novice-passive models, where NP→NP-I was more predictive than NP→NP-N. 

Novice passive supervision provided a usable, though weaker, proxy for expert observers. On novice demonstrations, NP→IP-N reached CC = 0.325 and NSS = 3.17, compared to CC = 0.365 and NSS = 43.60 for IP→IP-N. On intermediate demonstrations, this was reduced (NP→IP-I: CC = 0.426, NSS = 4.38 vs. IP→IP-I: CC = 0.446, NSS = 4.51). This suggests that crowd-sourced passive gaze can approximate expert-observer attention with limited loss, particularly when execution quality is high.

Collectively, these results demonstrate that the source and modality of gaze supervision materially shape learned perceptual strategies. Passive expert gaze encodes task-relevant structure that generalizes across performers, while scalable non-expert passive labels offer a viable, data-centric alternative for training surgical perception models, albeit with a predictable reduction in fidelity.

Overall, our findings have data-centric implications for how surgical gaze models should be trained in practice. Collecting operative expert gaze is the most expensive and least scalable modality, yet our results show that it is not strictly necessary as the primary supervision signal. Passive expert gaze captures a substantial portion of the operative attention structure and generalizes across performers, making it a high–value, lower–cost alternative for modeling expert perception. Moreover, the relatively small performance gap between expert-passive and novice-passive supervision on expert-observer targets suggests that large-scale crowd-sourced viewing data can potentially serve as an effective proxy for expert attention, particularly when demonstrations are of high technical quality. In contrast, novice active gaze, while abundant, is poorly aligned with expert perceptual strategies and may provide limited utility when used to train model for expert-oriented applications such as to inform surgical autonomy or decision making.

However, novice gaze data may still have utility for training models for education applications. In many educational settings, the system’s primary input will be novice performance rather than expert execution. Notably, novice-passive supervision remained a reasonable approximation of expert-observer attention on novice demonstrations. This indicates that even non-expert observers potentially tend to attend to many of the same diagnostically relevant regions when evaluating novice behavior. As a result, scalable novice-viewing data may be used to train perception models to support feedback and coaching pipelines that operate on novice motion, without requiring large number of expert annotation.

\section{Conclusion and Future Work}

This study provides a systematic, data-centric characterization of how the source of gaze supervision shapes learning-based surgical perception. Across tasks, active gaze exhibited skill-dependent organization consistent with prior RMIS findings, supporting the construct validity of the collection and processing pipeline. Passive gaze, while distinct from action-coupled attention, preserved substantial task-relevant structure, and models trained on passive labels recovered meaningful portions of operative attention, albeit with predictable degradation.

However, we do note several limitations in our dataset. First, novice recordings for Task D (needle driving) were scarce: only three novices attempted Task D, and none completed it due to session time constraints and insufficient proficiency. Second, the passive observation dataset did not include novice-source segments for Task D; therefore, we avoided making strong claims about novice passive gaze behavior on Task D, and we did not use Task D for novice–intermediate passive comparisons. Third, the intermediate group was small, and consisted of trained non-surgeons who were part of the study team rather than board-certified experts, which may have introduced an inherent expectancy bias, limiting generalization of our conclusions to actual clinical expertise levels. 

Future work will expand this dataset along three axes. First, we will recruit larger pools of clinically experienced surgeons to disentangle simulator proficiency from true clinical expertise and to validate whether the trends observed here persist at higher levels of skill. Second, we will strengthen coverage of difficult tasks, especially needle driving, through extended novice training and staged subtasks to ensure balanced passive-viewing schedules. Third, while this work focused on frame-wise saliency, we will investigate spatiotemporal and phase-conditioned gaze modeling to capture strategy-dependent dynamics such as error monitoring and recovery behavior.

Overall, these results suggest that expensive operative expert gaze is not the only viable supervision source for learning surgical perceptual priors. Passive viewing data, including scalable non-expert observation, can preserve substantial task-relevant structure and offers a practical pathway toward data-efficient, deployable gaze-guided perception models for surgical training and assistance.

\addtolength{\textheight}{-12cm}   





\section*{ACKNOWLEDGMENT}

This work made use of the High Performance Computing Resource in the Core Facility for Advanced Research Computing at Case Western Reserve University. 


\bibliographystyle{ieeetr}  
\bibliography{references}

\end{document}